\documentclass[11pt,a4paper]{article}
\usepackage[hyperref]{emnlp2020}
\usepackage{times}
\usepackage{latexsym}

\usepackage{microtype}
\usepackage{amsmath}
\usepackage{cleveref}
\usepackage{graphicx}
\usepackage{multirow}

\usepackage{booktabs,tabularx}
\usepackage{todonotes}
\def\parcite#1{\citep{#1}} % (Smith, 2012)
\def\perscite#1{\citet{#1}} % Smith (2012)
\def\transl#1#2{\texttt{#1}$\rightarrow$\texttt{#2}}
\def\pair#1#2{#1--#2}
\aclfinalcopy % Uncomment this line for the final submission
\title{CUNI Systems for the Unsupervised and Very Low Resource \\Translation Task in WMT20}

\author{Ivana Kvapil\'{i}kov\'{a}
        \qquad Tom Kocmi
        \qquad Ond{\v{r}}ej Bojar
		\\ \\
        Charles University, Faculty of Mathematics and Physics \\
        Institute of Formal and Applied Linguistics \\
        Malostransk{\'{e}} n{\'{a}}m{\v{e}}st{\'{\i}} 25, 118 00 Prague, Czech Republic \\
         {\tt <surname>@ufal.mff.cuni.cz}}

\date{}

\begin{document}
\maketitle
\begin{abstract}
This paper presents a description of CUNI systems submitted to the WMT20 task on unsupervised and very low-resource supervised machine translation between German and Upper Sorbian. We experimented with training on synthetic data and pre-training on a related language pair. In the fully unsupervised scenario, we achieved 25.5 and 23.7 BLEU translating from and into Upper Sorbian, respectively. Our low-resource systems relied on transfer learning from German--Czech parallel data and achieved 57.4 BLEU and 56.1 BLEU, which is an improvement of 10 BLEU points over the baseline trained only on the available small German--Upper Sorbian parallel corpus. 
\end{abstract}

\section{Introduction}

An extensive area of the machine translation (MT) research focuses on training translation systems without large parallel data resources \citep{artetxe2018nmt,artetxe2018smt,Artetxe2019effective,lample2018only,lample2018}. The WMT20 translation competition %\XXX{citaci az bude znama} 
presents a separate task on unsupervised and very low-resource supervised MT. %The language pair of interest is German (\texttt{de}) and Upper Sorbian (\texttt{hsb}). 

The organizers prepared a shared task to explore machine translation on a real-life example of a low-resource language pair of German (\texttt{de}) and Upper Sorbian (\texttt{hsb}). There are around 60k authentic parallel sentences available for this language pair which is not sufficient to train a high-quality MT system in a standard supervised way, and calls for unsupervised pre-training \citep{conneau2019pretraining}, data augmentation by synthetically produced sentences \citep{sennrich-bt} or transfer learning from different language pairs \citep{zoph2016,kocmi2018}.

The WMT20 shared task is divided into two tracks. In the unsupervised track, the participants are only allowed to use monolingual German and Upper Sorbian corpora to train their systems; the low-resource track permits the usage of auxiliary parallel corpora in other languages as well as a small parallel corpus in German--Upper Sorbian.

We participate in both tracks in both translation directions. Section~\ref{sec:unsup} describes our participation in the unsupervised track and \cref{sec:lowres} describes our systems from the low-resource track. Section~\ref{sec:lowres_trans} introduces transfer learning via Czech (\texttt{cs}) into our low-resource system. We conclude the paper in \cref{sec:concl}.

%\XXX{Pridat sekci Related Work?}

\section{Unsupervised MT}
\label{sec:unsup}
Unsupervised machine translation is the task of learning to translate without any parallel data resources at training time.  Both neural and phrase-based systems were proposed to solve the task \citep{lample2018}. In this work, we train several neural systems and compare the effects of different training approaches.

\begin{figure*}
    \centering
    \includegraphics[width=\textwidth]{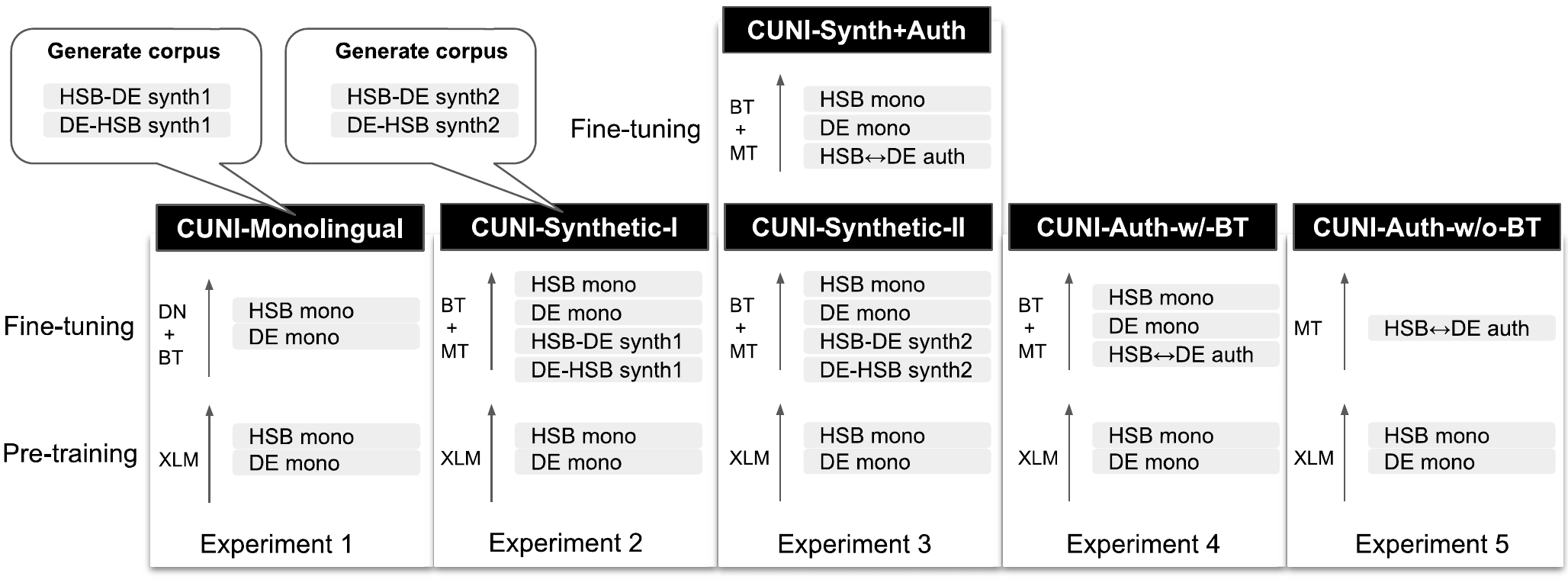}
    \caption{\textbf{An overview of selected CUNI systems.} Corpora are illustrated in gray boxes, system names in black boxes. Systems are trained with indicated training objectives: cross-lingual masked language modeling (XLM), de-noising (DN), online back-translation (BT), and standard machine translation objective (MT). Monolingual training sets \textit{DE~mono} and \textit{HSB~mono} were available for both WMT20 task tracks, the parallel training set \textit{HSB}$\leftrightarrow$\textit{DE~auth} was only allowed in the low-resource supervised track.
    }
    \label{fig:overview}
\end{figure*}

\subsection{Methodology}
\label{ssec:unsup_meth}
The key concepts of unsupervised NMT include a shared encoder, shared vocabulary and model initialization (pre-training). The training relies only on monolingual corpora and switches between de-noising, where the model learns to reconstruct corrupted sentences, and online back-translation, where the model first translates a batch of sentences and immediately trains itself on the generated sentence pairs, using the standard cross-entropy MT objective 
\citep{artetxe2018nmt,lample2018only}.
 
We use a 6-layer Transformer architecture for our unsupervised NMT models following the approach by \citet{conneau2019pretraining}. Both the encoder and the decoder are shared across languages. 

%We investigate the impact of pre-training only a bilingual \texttt{de-hsb} language model and a trilingual \texttt{de-cs-hsb} model and found that trilingual XLM pre-training significantly harms translation quality. Therefore, bilingual pre-training is used for all models described in this section. 

We first pre-train the encoder and the decoder separately on the task of cross-lingual masked language modelling (XLM) using monolingual data only \citep{conneau2019pretraining}. Subsequently, the initialized MT system (\textit{CUNI-Monolingual}) is trained using de-noising and online back-translation. We then use this system to translate our entire monolingual corpus and train a new system (\textit{CUNI-Synthetic-I}) from scratch on the two newly generated synthetic parallel corpora \textit{DE-HSB synth1} and \textit{HSB-DE synth1}. Finally, we use the new system to generate
\textit{DE-HSB synth2} and 
\textit{HSB-DE synth2}, and repeat the training to evaluate the effect of another back-translation round (\textit{CUNI-Synthetic-II}). 

All unsupervised systems are trained using the same BPE subword vocabulary \citep{sennrich} with 61k items generated using \texttt{fastBPE}.\footnote{\url{https://github.com/glample/fastBPE}} An overview of the systems and their training stages is given in \cref{fig:overview}.

\subsection{Data}
\label{ssec:unsup_data}
Our \texttt{de} training data comes from News Crawl; the \texttt{hsb} data was provided for WMT20 by the Sorbian Institute and the Witaj Sprachzentrum.\footnote{\url{http://www.statmt.org/wmt20/unsup_and_very_low_res/}} Most of the \texttt{hsb} data was of high quality but we fed the web-scraped corpus (\textit{web\textunderscore monolingual.hsb}) through a language identification tool \texttt{fastText}\footnote{\url{https://git hub.com/facebookresearch/fastText/}} to identify proper \texttt{hsb} sentences. All \texttt{de} data was also cleaned using this tool.

The final monolingual training corpora have 22.5M sentences (\textit{DE mono}) and 0.6M sentences (\textit{HSB mono}). Synthetic parallel corpora are generated from the monolingual data sets by coupling the sentences with their translation counterparts as described in \cref{ssec:unsup_meth}.

The parallel development (dev) and testing (dev test) data sets of 2k sentence pairs provided by WMT20 organizers are used for parameter tuning and model selection. The final evaluation is run on the blind test set \textit{newstest2020}.

\begin{table*}
\centering
\footnotesize
 \begin{tabular}{|l|c| c| c | c | c | c||c|} 
 \cline{3-8}
 \multicolumn{2}{c}{}&\multicolumn{5}{|c||}{\textbf{\textit{newstest2020}}} & \textbf{\textit{dev test set}}\\
 \hline
&\textbf{System Name} & \textbf{BLEU}  & \textbf{BLEU-cased}  & \textbf{TER} & \textbf{BEER 2.0} &  \textbf{CharacTER} & \textbf{BLEU} \\  [0.5ex] 
 \hline\hline

 \multirow{3}{*}{a} & \textbf{CUNI-Monolingual} & \textbf{23.7} & \textbf{23.4} & 0.606 &\textbf{0.530} & 0.559 & 23.4 \\ 
   \cline{2-8}
 &\textbf{CUNI-Synthetic-I} &23.4&23.2&0.617&0.531&0.575 & 22.2 \\
   \cline{2-8}
 &\textbf{CUNI-Synthetic-II*} & \textbf{23.7} & \textbf{23.4} & \textbf{0.618} & \textbf{0.530} & \textbf{0.563} & \textbf{23.7} \\
 
 \hline
 \hline
 \multirow{4}{*}{b} 
   &\textbf{CUNI-Supervised-Baseline} &43.7 & 43.2 & 0.439 & 0.670 & 0.382 & 38.7\\
 \cline{2-8}
  &\textbf{CUNI-Auth-w\textbackslash o-BT} & 51.6 & 51.2 & 0.362 & 0.710 & 0.332 & 48.3 \\
  \cline{2-8}
  &\textbf{CUNI-Auth-w\textbackslash -BT} & \textbf{54.3}  & \textbf{53.9} & \textbf{0.337} & \textbf{0.726} & \textbf{0.310} & \textbf{52.1} \\
  \cline{2-8} 
   &\textbf{CUNI-Synth+Auth*} & 53.8 & 53.4 & 0.343 & 0.721 & 0.315 & 50.5 \\

     \hline

\end{tabular}
\caption{Translation quality of the unsupervised (a)  and low-resource supervised (b) \texttt{hsb} $\rightarrow$ \texttt{de} systems on \textit{newstest2020} and the unofficial test set. The asterisk * indicates systems submitted into WMT20.} 
\label{tab:unsup_hsbde}
\end{table*}

\begin{table*}
\centering
\footnotesize
 \begin{tabular}{|l|c| c| c | c | c | c||c |} 
  \cline{3-8}
 \multicolumn{2}{c}{}&\multicolumn{5}{|c||}{\textbf{\textit{newstest2020}}}& \textbf{\textit{dev test set}}\\
 \hline
 & \textbf{System Name} & \textbf{BLEU}  & \textbf{BLEU-cased}  & \textbf{TER} & \textbf{BEER 2.0} &  \textbf{CharacTER} & \textbf{BLEU}   \\  [0.5ex] 
 \hline\hline

 \multirow{3}{*}{a} & \textbf{CUNI-Monolingual} & 21.7 & 21.2 & 0.670 & 0.497 & 0.557  & 20.4 \\ 
   \cline{2-8}
 & \textbf{CUNI-Synthetic-I} & 24.9 & 24.5 & 0.599 & 0.535 & 0.521 & 25.1 \\
 \cline{2-8}
 & \textbf{CUNI-Synthetic-II*} & \textbf{25.5} & \textbf{25.0} & \textbf{0.592} & \textbf{0.540} & \textbf{0.516} & \textbf{25.3} \\

 \hline
 \hline
 \multirow{4}{*}{b}
   &\textbf{CUNI-Supervised-Baseline} & 40.8 & 40.3 & 0.452 & 0.655 & 0.373 & 38.3\\
  \cline{2-8}
  &\textbf{CUNI-Auth-w\textbackslash o-BT}  & 47.5 & 47.1 & 0.390 & 0.689 & 0.336 & 47.1 
\\
  \cline{2-8}
  &\textbf{CUNI-Auth-w\textbackslash -BT} & \textbf{52.3} & \textbf{51.8} & \textbf{0.350} & \textbf{0.718} & \textbf{0.301} & \textbf{52.4}\\
  \cline{2-8} 
 & \textbf{CUNI-Synth+Auth*} & 50.6 & 50.1 & 0.368 & 0.703 & 0.326 & 50.4 \\
     \hline
 \hline

\end{tabular}
\caption{Translation quality of the unsupervised (a) and low-resource supervised (b) \texttt{de} $\rightarrow$ \texttt{hsb} systems on \textit{newstest2020} and the unofficial test set. The asterisk * indicates systems submitted into WMT20.}
\label{tab:unsup_dehsb}
\end{table*}

\subsection{Results}

%We selected the model with the highest scores on the development data sets. 
The resulting scores measured on the blind \textit{newstest2020} are listed in \cref{tab:unsup_hsbde} and \cref{tab:unsup_dehsb}. The translation quality metrics BLEU \citep{bleu}, TER \citep{ter_Snover06astudy}, BEER \citep{beer_stanojevic-simaan-2014-fitting} and CharacTER \citep{character_wang-etal-2016-character} provide consistent results. The best quality is reached when using synthetic corpora from the second back-translation iteration, although the second round adds only a slight improvement. A similar observation is made by \perscite{hoang2018iterative} who show that the second round of back-translation does not enhance the system performance as much as the first round. Additionally, the third round does not produce any significant gains.
%\tk{moje intuice v tomhle je, ze tim jak nepridavas novou informaci a pouze zlepsujes preklad reference synteetickych dat, tak uz nemuzes ziskat lepsi vysledky a naopak po nekolika iteracich clovek muze pretrenovat model na synthetyckych datech a vyrazne zhorsit kvalitu prekladu}

When training on synthetic parallel corpora, it is still beneficial to perform back-translation on-the-fly \citep{artetxe2018nmt} whereby new training instances of increasing quality are generated in every training step. This method adds 1 - 2 BLEU points to the final score as compared to training only on sentence pairs from the two synthetic corpora so we use it in all our unsupervised systems.

We used the \texttt{XLM}\footnote{\url{https://github.com/facebookresearch/XLM}} toolkit for running the experiments. Language model pre-training took 4 days on 4 GPUs\footnote{GeForce GTX 1080, 11GB of RAM}. The translation models were trained on 1 GPU\footnote{Quadro P5000, 16GB of RAM} with 8-step gradient accumulation to reach an effective batch size of 8 $\times$ 3400 tokens. We used the Adam \citep{adam} optimizer with inverse square root decay ($\beta_1 =0.9$, $\beta_2 = 0.98$, $lr = 0.0001$) and greedy decoding.

\begin{table*}
\centering
\footnotesize
 \begin{tabular}{|c| c| c | c | c | c|} 
 \hline
\textbf{System Name} & \textbf{BLEU}  & \textbf{BLEU-cased}  & \textbf{TER} & \textbf{BEER 2.0} &  \textbf{CharacTER}  \\  [0.5ex] 
 \hline\hline
 \textbf{Helsinki-NLP}  	 & 60.0 & 59.6 & 0.286 & 0.761 & 0.267 \\
% Scheduled multi-task learning with subword sampling. 24/24r: with monolingual noise task	yes	24+24r ensemble
\textbf{NRC-CNRC} & 59.2 & 58.9 & 0.290 & 0.759 & 0.268 \\
% (Details)	NRC-CNRC National Research Council Canada	Low-resource supervised. Transformer, transfer learning, BPE-dropout, backtranslation. Ensemble.	yes
\textbf{SJTU-NICT} & 58.9 & 58.5 & 0.296 & 0.754 & 0.274 \\
% (Details)	charlee SJTU-NICT	Low-resource semi-supervised NMT, Cross-lingual Language modeling (CMLM), Translation Language modeling (TLM), Data-dependent Gaussian Prior Objective (D2GPo), BT-BLEU Co-filter based Self Training (CFST), UNMT as Pretraining	yes	ensemble 2, fix quotes
 \textbf{CUNI-Transfer} & 57.4 & 56.9 & 0.307 & 0.746 & 0.285 \\
 \textbf{Bilingual only} & 47.8 & 47.4 & 0.394 & 0.695 & 0.356 \\
 \hline
\end{tabular}
\caption{Translation quality of \texttt{hsb} $\rightarrow$ \texttt{de} systems on newstest2020.}
\label{tab:lowres_hsbde}
\end{table*}
\begin{table*}
\centering
\footnotesize
 \begin{tabular}{|c| c| c | c | c | c|} 
 \hline
\textbf{System Name} & \textbf{BLEU}  & \textbf{BLEU-cased}  & \textbf{TER} & \textbf{BEER 2.0} &  \textbf{CharacTER}  \\  [0.5ex] 
 \hline\hline
 \textbf{SJTU-NICT} & 61.1 & 60.7 & 0.283 & 0.759 & 0.250 \\
%  Low-resource semi-supervised NMT, Cross-lingual Language modeling (CMLM), Translation Language modeling (TLM), Data-dependent Gaussian Prior Objective (D2GPo), BT-BLEU Co-filter based Self Training (CFST), UNMT as Pretraining	yes	ensemble 2, L2R + R2L reranking
\textbf{Helsinki-NLP} & 58.4 & 57.9 & 0.297 & 0.755 & 0.255 \\
% (Details)	yvesscherrer University of Helsinki	Low-resource supervised. Low-resource supervised. Scheduled multi-task learning with subword sampling. 25/25r: no monolingual tasks	yes	25+25r ensemble
\textbf{NRC-CNRC} & 57.7 & 57.3 & 0.300 & 0.754 & 0.255 \\
% (Details)	NRC-CNRC National Research Council Canada	Low-resource supervised. Transformer, transfer learning, BPE-dropout, backtranslation. Ensemble.	yes	
 \textbf{CUNI-Transfer} & 56.1 & 55.5 & 0.315 & 0.743 & 0.265 \\ 
 \textbf{Bilingual only} & 46.8 & 46.4 & 0.389 & 0.692 & 0.335 \\
 \hline
\end{tabular}
\caption{Translation quality of \texttt{de} $\rightarrow$ \texttt{hsb} systems on newstest2020.}
\label{tab:lowres_transf_dehsb}
\end{table*}

\section{Very Low-Resource Supervised MT}
\label{sec:lowres}
\subsection{Methodology}
%We used two system architectures and training approaches in our submissions to the low-resource supervised track: pre-training on synthetic data and transfer learning via Czech. 

Our systems introduced in this section have the same model architecture as described in \cref{sec:unsup}, but now we allow the usage of authentic parallel data. We pre-train a bilingual XLM model and fine-tune with either only authentic parallel data (\textit{CUNI-Auth-w\textbackslash o-BT}) or both parallel and monolingual data, using a combination of standard MT training and online back-translation (\textit{CUNI-Auth-w\textbackslash-BT}). Finally, we utilize the trained model \textit{CUNI-Synthetic-II} from \cref{sec:unsup} and fine-tune it on the authentic parallel corpus, again using standard supervised training as well as online back-translation (\textit{CUNI-Synth+Authentic)}. 

All systems are trained with the same BPE subword vocabulary of 61k items.

\subsection{Data}
\label{ssec:lowres_data}
In addition to the data described in \cref{ssec:unsup_data}, we used the authentic parallel corpus of 60k sentence pairs provided by Witaj Sprachzentrum mostly from the legal and general domain. 

\subsection{Results}
The resulting scores are listed in the second part of \cref{tab:unsup_hsbde} and \cref{tab:unsup_dehsb}. We compare system performance against a supervised baseline which is a vanilla NMT model trained only on the small parallel train set of 60k sentences, without any pre-training or data augmentation. 

Our best system gains 11.5 BLEU over this baseline, utilizing the larger monolingual corpora for XLM pre-training and online back-translation.  Fine-tuning one of the trained unsupervised systems on parallel data leads to a lower gain of $\sim$10 BLEU points over the baseline.

The translation models were trained on 1 GPU\footnote{GeForce GTX 1080 Ti, 11GB of RAM} with 8-step gradient accumulation to reach an effective batch size of 8 $\times$ 1600 tokens. Other training details are equivalent to \cref{ssec:unsup_meth}.% The translation quality was still improving after 2 days of training. 

\section{Very Low-Resource Supervised MT with Transfer Learning} 
\label{sec:lowres_trans}

One of the main approaches to improving performance under low-resource conditions is transferring knowledge from different high-resource language pairs \parcite{zoph2016transferLowResource,kocmi2018}. This section describes the unmodified strategy for transfer learning as presented by \perscite{kocmi2018}, using German--Czech as the parent language pair. Since we do not modify the approach nor tune hyperparameters of the NMT model, we consider our system a transfer learning baseline for low-resource supervised machine translation. %We submitted our system with the name \textit{CUNI-Transfer}.

\subsection{Methodology}

\perscite{kocmi2018} proposed an approach to fine-tune a low-resource language pair (called ``child'') from a pre-trained high-resource language pair (called ``parent'') model. The method has only one restriction and that is a shared subword vocabulary generated from the corpora of both the child and the parent.
The training procedure is as follows: first train an NMT model on the parent parallel corpus until it converges, then replace the training data with the child corpus. 

We use the Tensor2Tensor framework \parcite{tensor2tensor} for our transfer learning baseline and model parameters ``Transformer-big'' as described in \parcite{tensor2tensor}. Our shared vocabulary has 32k wordpiece tokens. We use the Adafactor \parcite{shazeer2018adafactor} optimizer and a reverse square root decay with 16 000 warm-up steps. For the inference, we use beam search of size 8 and alpha 0.8.

\subsection{Data}

In addition to the data described in \cref{ssec:lowres_data}, we used the \texttt{cs}-\texttt{de} parallel corpora available at the OPUS\footnote{\url{http://opus.nlpl.eu/}} website: OpenSubtitles, MultiParaCrawl, Europarl, EUBookshop, DGT, EMEA and JRC. The \texttt{cs}-\texttt{de} corpus  has 21.4M sentence pairs after cleaning with the \texttt{fastText} language identification tool.

\subsection{Results}
We compare the results of our transfer learning baseline called \textit{CUNI-Transfer} with three top performing systems of WMT20. These systems use state-of-the-art techniques such as BPE-dropout, ensembling of models, cross-lingual language modelling, filtering of training data and hyperparameter tuning. Additionally, we also include results for a system we trained without any modifications solely on bilingual parallel data (\textit{Bilingual only}).\footnote{The model \textit{Bilingual only} is trained on the same data as \textit{CUNI-Supervised-Baseline} but uses a different architecture and decoding parameters.}

The results in \cref{tab:lowres_transf_dehsb} show that training solely on German--Upper Sorbian parallel data leads to a performance of 47.8 BLEU for \transl{de}{hsb} and  46.7 BLEU for \transl{hsb}{de}. When using transfer learning with a \pair{Czech}{German} parent, the performance increases by roughly 10 BLEU points to 57.4 and 56.1 BLEU. As demonstrated by the winning system, the performance can be further boosted using additional techniques and approaches to 60.0 and 61.1 BLEU. This shows that transfer learning plays an important role in the low-resource scenario.

\section{Conclusion}
\label{sec:concl}

We participated in the unsupervised and low-resource supervised translation task of WMT20. %Our best unsupervised model was submitted under the name \textit{CUNI-Synthetic-II}

In the fully unsupervised scenario, the best scores of 25.5 (\texttt{hsb}$\rightarrow$\texttt{de}) and 23.7 (\texttt{de}$\rightarrow$\texttt{hsb}) were achieved using cross-lingual language model pre-training (XLM) and training on synthetic data produced by NMT models from earlier two iterations. We submitted this system under the name \textit{CUNI-Synthetic-II}.

In the low-resource supervised scenario, the best scores of 57.4 (\texttt{hsb}$\rightarrow$\texttt{de}) and 56.1 (\texttt{de}$\rightarrow$\texttt{hsb}) were achieved by pre-training on a large German--Czech parallel corpus and fine-tuning on the available German--Upper Sorbian parallel corpus. We submitted this system under the name \textit{CUNI-Transfer}.

We showed that transfer learning plays an important role in the low-resource scenario, bringing an improvement of $\sim$10 BLEU points over a vanilla supervised MT model trained on the small parallel data only. Additional techniques used by other competing teams yield further improvements of up to 4 BLEU over our transfer learning baseline.

%\tk{do zaveru by se hodilo nejake shrnuti co je uzitecne a co spis nefungovalo, aby to bylo selfcontained}

\section*{Acknowledgments}

This study was supported in parts by the grants
19-26934X % Ivana-Expro
and 18-24210S % Ondrej a Tom GACR obyc
of the Czech Science Foundation, SVV~260~575 and GAUK 1050119 of the Charles University.
%, and 825303 (Bergamot) of the European Union.
This work has been using language resources and tools stored and distributed by the LINDAT/CLARIN project of the Ministry of Education, Youth and Sports of the Czech Republic (LM2018101).

\bibliography{anthology,biblio}
\bibliographystyle{acl_natbib}

\appendix

\end{document}